\documentclass[dvipsnames,format=sigconf,anonymous=false,review=false]{acmart}
\usepackage{annotate-equations}
\usepackage{algorithm}
\usepackage{algpseudocode}
\usepackage{enumitem}

\AtBeginDocument{%
  \providecommand\BibTeX{{%
    \normalfont B\kern-0.5em{\scshape i\kern-0.25em b}\kern-0.8em\TeX}}}

\setcopyright{acmcopyright}
\copyrightyear{2024}
\acmYear{2024}
\acmDOI{10.1145/1122445.1122456}

\acmConference[GECCO 2024]{Genetic
and Evolutionary Computation Conference Companion}{2024}{Melbourne}

\begin{document}

\title{On the robustness of lexicase selection to contradictory objectives}
\author{Shakiba Shahbandegan}
 \email{shahban1@msu.edu}
 \orcid{0000-0002-3670-630X}
 \affiliation{%
   \institution{Michigan State University}
   \city{East Lansing}
   \state{Michigan}
   \country{USA}
 }

\author{Emily Dolson}
 \email{dolsonem@msu.edu}
 \orcid{0000-0001-8616-4898}
 \affiliation{%
   \institution{Michigan State University}
   \city{East Lansing}
   \state{Michigan}
   \country{USA}
 }

\renewcommand{\shortauthors}{Shahbandegan and Dolson}

\begin{abstract}
Lexicase and $\epsilon$-lexicase selection are state of the art parent selection techniques for problems featuring multiple selection criteria. Originally, lexicase selection was developed for cases where these selection criteria are unlikely to be in conflict with each other, but preliminary work suggests it is also a highly effective many-objective optimization algorithm. However, to predict whether these results generalize, we must understand lexicase selection's performance on contradictory objectives. Prior work has shown mixed results on this question. Here, we develop theory identifying circumstances under which lexicase selection will succeed or fail to find a Pareto-optimal solution. To make this analysis tractable, we restrict our investigation to a theoretical problem with maximally contradictory objectives. Ultimately, we find that lexicase and $\epsilon$-lexicase selection each have a region of parameter space where they are incapable of optimizing contradictory objectives. Outside of this region, however, they perform well despite the presence of contradictory objectives. Based on these findings, we propose theoretically-backed guidelines for parameter choice. Additionally, we identify other properties that may affect whether a many-objective optimization problem is a good fit for lexicase or $\epsilon$-lexicase selection.
\end{abstract}




\maketitle

\section{Introduction}

In recent years, lexicase selection and its variants such as $\epsilon$-lexicase selection have emerged as state of the art approaches to parent selection in many types of evolutionary computation \citep{spector_assesment_2012, lacavaEpsilonLexicaseSelectionRegression2016, helmuth_benchmarking_2020}. These algorithms solve problems where fitness depends on multiple distinct criteria. Initially designed for genetic programming applications, lexicase selection has proven to successfully tackle a wide range of problem domains such as optimizing neural networks \cite{ding2022optimizing}, learning classifier systems \cite{10.1145/3321707.3321828}, evolutionary robotics \cite{10.1162/artl_a_00374} and feature selection \cite{10.1007/978-3-319-55696-3_6}.
One particularly notable area where lexicase selection has shown promise is many-objective optimization \citep{lacavaProbabilisticMultiObjectiveAnalysis2018, lacavaAnalysislexicaseSelection2018a}. Findings from previous studies prompt a broad question: Is lexicase selection a reliable general-purpose evolutionary many-objective optimization algorithm, or are there specific many-objective problems for which it may not be well-suited? 

\begin{figure}
    \centering
    \includegraphics[width=.7\linewidth]{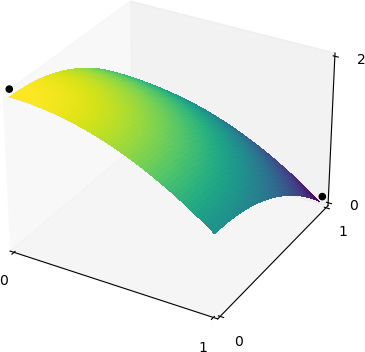}
    \caption{The full Pareto front vs. the solution set maintained by lexicase selection. Surface shows the set of points that fall along the Pareto front for a hypothetical 3-objective problem. The two points in black represent points that lexicase selection would maintain if the entire theoretical Pareto front were in the population. A third point could be maintained if the points with the highest x and  y values were different.}
    \label{fig:pareto_cartoon}
\end{figure}

Note that we limit this analysis to many- and massive- objective problems, because lexicase selection is obviously a bad choice for most classic two and three objective problems. While all solutions maintained by lexicase selection are Pareto-optimal, it only maintains a small fraction of the full Pareto front (see Figure \ref{fig:pareto_cartoon}) \citep{lacavaProbabilisticMultiObjectiveAnalysis2018}. For many traditional multi-objective optimization problems, this property renders lexicase selection inappropriate. 
For example, a classic multi-objective optimization problem is trying to choose a car that maximizes quality and minimizes cost. Most people faced with this problem want a car with intermediate quality and intermediate cost - i.e. a solution from the middle of the Pareto front. Lexicase selection, on the other hand, would only give the option of choosing either the most expensive car or the lowest quality car. 

However, as the number of objectives increases, the importance of maintaining the entire Pareto front becomes less clear \citep{lacavaProbabilisticMultiObjectiveAnalysis2018}. In a problem with, for example, 100 objectives, a population can only reasonably contain a small fraction of the full Pareto front. In such situations, does it matter if the population only contains solutions that are top performers on at least one of the 100 objectives? There are two possible reasons that it might: 1) the user may want a perfect compromise among all objectives, and 2) maintaining ``compromise'' objectives could aid the evolutionary process. We suspect that instances of the former are relatively uncommon with such high numbers of objectives, although in such cases lexicase selection is clearly the wrong algorithm to choose. The latter is an empirical question, worthy of further research. 

Nevertheless, empirical evidence thus far suggests that lexicase and $\epsilon$-lexicase selection perform well on many-objective optimization problems. For example, on DTLZ problems (specifically DTLZ1, DTLZ2, DTLZ3, and DTLZ4) with 5 or more objectives, $\epsilon$-lexicase outperformed NSGA-II \citep{lacavaAnalysislexicaseSelection2018a}. $\epsilon$-lexicase also performed well on a larger suite of problems with five or more objectives \citep{lacavaProbabilisticMultiObjectiveAnalysis2018}. However, anecdotally, in other cases lexicase selection has failed to achieve performance comparable to purpose-built multi-objective algorithms like NSGA-II. 

What causes lexicase and $\epsilon$-lexicase to perform well on some many-objective problems but poorly on others? An obvious explanation is that the relationships between the objectives are a critical factor. 
Lexicase selection was developed for genetic programming, in which fitness criteria generally correspond to individual test cases that evolved code must pass. Test cases are different from traditional objectives in a few important ways \citep{helmuthGeneralProgramSynthesis}. Most notably, it is uncommon that improving performance on one test case would make a solution perform worse on a different test cases. In contrast, multi-objective optimization problems are usually assumed to involve goals with inherent trade-offs. In other words, optimizing one objective will necessarily lead to the deterioration of performance on the other objectives. 

Do such contradictory objectives inhibit lexicase selection's ability to solve many-objective problems? Clearly, based on the DTLZ results \citep{lacavaAnalysislexicaseSelection2018a}, $\epsilon$-lexicase selection at least can tolerate objectives that are not perfectly aligned with each other. However, on a problem with a vast number of contradictory objectives, non-dominated sorting out-performed lexicase selection \citep{hernandez_suite_2022}. Subsequent research suggested that lexicase selection's performance is further harmed when objectives have more intense trade-offs with each other \citep{shahbandegan_untangling_2022}. 

Any pair of objectives can be categorized along a continuum containing the following three broad categories: 1) \textbf{aligned}, meaning any solution that improves on the first objective also improves on the second objective, 2) \textbf{orthogonal}, meaning that performance on the first objective has no relationship to performance on the second objective, or 3) \textbf{contradictory}, meaning improving performance on the first objective decreases performance on the other and visa versa. 
We hypothesize that lexicase selection's ability to solve a many-objective optimization problem depends on the number of groups of orthogonal and contradictory objectives.

Here, we explore this hypothesis via mathematical modeling of lexicase selection and $\epsilon$-lexicase selection in a many-objective optimization problem with contradictory objectives. We find that, given an appropriate population size, lexicase selection is surprisingly robust to the presence of contradictory objectives, while $\epsilon$ lexicase selection appears to be more impacted by them. Our results inform algorithm selection decisions and aid in the identification of appropriate parameter values for lexicase and $\epsilon$-lexicase selection on many-objective problems.

\section{Background}

\subsection{Lexicase Selection}
Lexicase is a parent selection algorithm in evolutionary computation. It is designed for selecting candidate solutions for reproduction based on multiple fitness criteria/test cases \citep{spector_assesment_2012}. To select a parent, lexicase selection iterates through a set of fitness criteria in a randomly shuffled order. At each step, all but the individuals with the highest fitness on the current test case are eliminated. This process continues until only one individual is left; this individual is selected. If there are no test cases left and multiple candidates remain, an individual will be selected randomly. 

\subsection{$\epsilon$-Lexicase Selection}

Since lexicase selection keeps only individuals with the highest fitness at each step, it is often unsuitable for problems with continuous fitness criteria. In such contexts, individuals are unlikely to have a fitness exactly equal to the fitness of the elite individual in the selection pool. Consequently, ties rarely occur and the algorithm decays to elitist selection on the fitness criteria. To overcome this problem, $\epsilon$-lexicase selection was proposed \cite{lacavaEpsilonLexicaseSelectionRegression2016}. In $\epsilon$-lexicase selection, individuals within an $\epsilon$ threshold of the fittest individual are kept in the selection pool at each step. The value of $\epsilon$ can be a constant or change automatically throughout the run of evolution. When $\epsilon$ changes automatically, a common approach is to define it for each objective using median absolute deviation: \citep{lacavaEpsilonLexicaseSelectionRegression2016}:
\begin{equation}
    \epsilon(j) = median(|O_j(z_i) - median(O_j(z_i))|) \forall z_i \in Z
\label{eq:medianepsilon}
\end{equation}
Where $O_j(z_i)$ is the value of objective $O_j$ for individual $z_i$ in population $Z$. Note that standard lexicase selection is equivalent to $\epsilon$-lexicase in which $\epsilon=0$. Depending on the problem at hand, there are three common ways to implement $\epsilon$-lexicase selection \cite{lacavaProbabilisticMultiObjectiveAnalysis2018}:
\begin{itemize}[itemindent=*, leftmargin=*]
    \item \textbf{Static:} In this version, $\epsilon$ is calculated once every generation for each case across the population. Individuals will have a pass/fail fitness depending on whether they are within the $\epsilon$ threshold of the best fitness over the entire population. 
    \item \textbf{Semi-dynamic:} Unlike static $\epsilon$-lexicase, in this version the pass condition is defined relative to the highest fitness among the current pool, while $\epsilon$ is calculated once every generation.
    \item \textbf{Dynamic:} In this variant, both $\epsilon$ and the pass condition are defined relative to the current selection pool.
\end{itemize}

We use semi-dynamic $\epsilon$-lexicase selection in our experiments. We analyze the impact of using constant values for $\epsilon$ as well as the impact of selecting it according to Equation \ref{eq:medianepsilon}. Pseudo-code for $\epsilon$-lexicase selection is shown in Algorithm 1. 

\begin{algorithm}
\caption{The semi-dynamic $\epsilon$-lexicase selection algorithm }\label{alg:cap}
\begin{algorithmic}
\Require Z - a vector of $S$ candidate solutions with scores on $D$ fitness criteria 
\Ensure A single candidate solution from P that should reproduce
\State $F \gets 1...D$ \Comment{Set of fitness criteria to consider}
\State $C \gets Z$ \Comment{Set of candidate solutions to consider}
\State $\epsilon \gets \epsilon_0$ \Comment{Assign constant value for epsilon}

\While{$|P| > 1$ \& $|F| > 0$}
\State curr $\gets$ random criterion in $F$
\State best $\gets$ -1 \Comment{Assumes criteria are being maximized}
\For{solution : $C$}
\If{solution[curr] > best - $\epsilon$}
\State best $\gets$ solution[curr]
\EndIf
\EndFor
\State $C \gets$ all members of $C$ with score best on criterion curr
\EndWhile

\hspace{-2em}\Return{A random candidate solution in $C$}
\end{algorithmic}
\end{algorithm}

\subsection{Related work}

\subsubsection{Mathematical analyses} 
\label{sec:prior_math}

Previously, La Cava et. al identified that a hugely influential value in predicting the outcome of lexicase selection is $P_{lex}$, the probability that a given solution in the population is selected on any given round of lexicase selection \citep{lacavaProbabilisticMultiObjectiveAnalysis2018}. $P_{lex}$ depends on $i$, the solution we are calculating the selection probability for, and $Z$, the current population (note that Equation \ref{eq:plex} also takes the current set of objectives as an argument, but this input is only necessary for recursive steps). $P_{lex}$ can be calculated with the following equation, adapted from \citep{lacavaProbabilisticMultiObjectiveAnalysis2018}:

\begin{align}
    P_{lex}(i | Z, N) =  \begin{cases}
    1 & if | Z | = 1 \\
    1/|Z| & if |N|=0 \\
    \frac{\sum\limits_{j = 0}^{|N|} P_{lex}(i, \{z \in Z | z \textit{ elite on } N_j \},  \{n_i \in N | i != j \})}{|N|} & else
    \end{cases}
\label{eq:plex}
\end{align}

where being ``elite'' on an objective is defined as having a fitness within $\epsilon$ threshold of the highest score on that objective within the population. More formally:

\begin{definition}
    A candidate solution, $z_i$, is elite on an objective $j$, within a population $Z$ if and only if $O_j(z_i) + \epsilon \geq O_j(z_k), \forall z_k \in Z$.
    \label{def:elite}
\end{definition}

Calculating $P_{lex}$ is $NP$-Hard, which has historically posed an obstacle to the type of analyses we carry out here \citep{dolsonCalculatingLexicaseSelection2023}. However, using the optimizations described in \citep{dolsonCalculatingLexicaseSelection2023}, we are able to calculate it in a tractable amount of time as long as either population size ($S$) or dimension ($D$) is not too high.

Knowing the probability of a given solution being selected is valuable, but in many contexts it does not tell the whole story. Often, a large network of neutral genotypes must be traversed before discovering one with a novel value on an objective function \cite{neutralspaces}.
In these cases, the relevant question is whether a sub-population of solutions with a given profile of objective function performance can survive long enough to traverse this network and discover a novel, improved objective function performance. 
Moreover, in the context of multi-objective optimization, it is generally desirable to maintain the entire Pareto front (to the extent that it fits within the population). 
In these cases, we need to know the probability that solutions with a given objective function performance profile will survive until the end of the run.
Dolson et. al described this value as $P_{survival}$, the probability that a solution will survive for some number of generations, $G$, assuming the population does not change.
This value is related to $P_{lex}$ and can be calculated with the following equation (adapted from \citep{dolson_ecological_2018}):

\begin{equation}
    P_{survival}(i, S, G, Z) = (1 - (1 - P_{lex}(i, Z))^S)^G
\label{eq:p_survival}
\end{equation}

Dolson et. al found that as $G$ and $S$ increase, this equation begins to approximate a step function \citep{dolson_ecological_2018}. This observation implies that when $G$ and $S$ are large enough, we can have a high degree of confidence about whether a given candidate solution will still be in the population after $G$ generations from the time it arises. Based on this fact, we will make the simplifying assumption going forward that all solutions with $P_{survival}$ greater than some threshold value $t$ are guaranteed to survive for $G$ generations, and all other solutions are guaranteed to go extinct within $G$ generations. Due to the step-function-like properties of Equation \ref{eq:p_survival}, as long as $G$ and $S$ are relatively large, the choice of $t$ should be relatively inconsequential. 

\subsubsection{Experimental Analyses} 

Hernandez et al. developed a suite of diagnostic fitness landscapes designed to identify the strengths and weaknesses of evolutionary algorithms under different conditions \cite{hernandez_suite_2022}. A diagnostic fitness landscape is a mapping from genotype to phenotype to fitness. Each diagnostic landscape is crafted carefully to assess an algorithm's strengths and weaknesses. The diagnostic most relevant to our work here is the ``contradictory objectives'' diagnostic, which gives  insight into an algorithm's ability to maintain diversity and find global optima on problems with multiple contradictory objectives. Hernandez et. al found that lexicase selection performed well on this diagnostic, but not nearly as well as non-dominated sorting. This finding lends credence to the idea that lexicase selection may struggle more with contradictory objectives than a purpose-built multi-objective optimization algorithm. 

However, Hernandez et. al \cite{hernandez_suite_2022} made an important observation: the prior mathematical results in \citep{lacavaProbabilisticMultiObjectiveAnalysis2018, dolson_ecological_2018} suggest that the performance of lexicase selection on this problem is entirely driven by population size. We will show later that they were correct in this assessment.

\section{Methods}

\subsection{Fitness Function}

To explore the constraints on lexicase selection's performance under contradictory objectives, we designed a fitness function with objectives that maximally oppose each other. This fitness function is closely inspired by the contradictory objectives diagnostic from the diagnostic suite proposed in \citep{hernandez_suite_2022}, specifically the antagonistic version described in \citep{shahbandegan_untangling_2022}.

Each candidate solution is represented by a ``genotype'' composed of $D$ integers between 0 and an upper limit $L$.
This genotype is then translated into a vector of scores on each of the objectives according to the following equation:

\begin{equation}
    O_i = v_i - \sum\limits_{\substack{j=0 \\ j \neq i}}^{D} v_j
    \label{eq:fitness_function}
\end{equation}

Consequently, increasing the value at any position in the genotype increases the score for the corresponding objective. However, it also decreases the score for every other objective. Consequently, the highest possible score for each objective is $L$ (achieved when the genotype value at that position is $L$ and all other values are 0). Because lexicase selection can only select solutions that are elite on at least one objective, these are the only Pareto-optimal solutions that lexicase selection is capable of selecting. As discussed in the introduction, this set of solutions lacks Pareto-optimal solutions with non-zero values in multiple positions. 


Note that, while our genotypes are intentionally simple, in our model they serve as abstractions of more complicated genotypic representations. For example, taking inspiration from \citep{cullyRobotsThatCan2015a}, we could imagine that we are evolving gaits for a six-legged robot. In this case, the genotype would be the code/neural network/other representation for the robot control algorithm. We could translate this complex genotype into a simple vector of behavioral descriptors: 6 numbers indicating the percentage of time each leg was touching the ground. We could then create various objectives associated with specific legs touching the ground for specific percentages of time. Some of these objectives would be inherently contradictory, because they would place different requirements on the percentage of time a given leg must be touching the ground. 

Importantly, for a more complex evolutionary scenario like this one, changes that will be reflected in the objective scores likely involve traversing a substantial sequence of neutral mutations. Consequently, a set of solutions with a given objective score profile must remain in the population for a substantial amount of time (represented here by the parameter $G$) before successfully producing an offspring with different objective scores. In large genotypic spaces, this process will likely take many generations \citep{fortunaGenotypephenotypeMapEvolving2017, neutralspaces}. $G$, then, should be the average number of generations required to cross a neutral genotypic space from one objective score profile to another. Thus, it is a property inherent to the problem being solved. Here, we make the simplifying assumption that it is a consistent value, which makes the analyses in this paper tractable. While it is unlikely that real-world problems have any single value of $G$ that applies to all neutral regions of their fitness landscapes, there is no reason to think that holding $G$ constant should affect lexicase selection's response to contradictory objectives. Thus, we feel that this simplifying assumption is acceptable.

\begin{table*}[htbp]
    \centering
    \begin{tabular}{lp{0.7\linewidth}}
        \toprule
        \textbf{Parameter} & \textbf{Description} \\
        \midrule
        Population size (S) & The number of selection events that will occur per generation. The population size plays a key role in determining the diversity and variability of solutions that can be maintained in the population. \\
        Dimensionality (D) & The number of objectives being used for the problem. \\
        Value Limit (L) & The maximum value that an objective score is allowed to have. \\
        Number of generations (G) & The number of generations that a genotype needs to survive for in order to traverse the neutral space between different objective scores. \\
        mutation rate ($\mu$) & The probability that a mutation occurs that changes a value in the genotype (once $G$ generations have elapsed). When a mutation occurs, the value can either increase or decrease by one unit. This mutation rate is per-genome (for each genotype, the probability that any value in the genome is mutated.) \\
        Threshold value (t) & The parameter indicating which solutions we expect to survive for $G$ generations based on the value of $P_{survival}$. Here, we use $t = .5$.\\
        \bottomrule
    \end{tabular}
    \caption{Description of parameters}
    \label{tab:param_table}
\end{table*}

\subsection{Stochastic Modelling}

To examine the behavior of lexicase and $\epsilon$-lexicase selection on realistically-sized instances of our problem, we turn to stochastic modelling. We use the fitness function described in the previous section to model scores that might be produced by a more complex underlying genetic representation. Our model involves multiple parameters that affect the performance of lexicase and $\epsilon$-lexicase selection on a many-objective optimization problem. Some of these parameters are under the control of users of these algorithms, while others are properties of the problem being solved. All parameters are described in \ref{tab:param_table}.
The probability that lexicase selection will fail to find a Pareto-optimal solution within the allotted time can be calculated recursively by determining the probability of arriving at each intermediate population state. Let $P_i$ be the probability of discovering objective score profile $i$. $P_i$ can be calculated by summing up the probabilities of arriving at all mutationally adjacent score profiles multiplied by the probability of discovering score profile $i$ from each of these profiles:

\vspace{1.5em}

\begin{equation}
    \eqnmarkbox[violet]{Pi}{P_{i}} = \sum_{j \in \eqnmarkbox[orange]{ai}{a(i)}} \eqnmarkbox[blue]{Pj}{P_j} * \eqnmarkbox[red]{Pji}{P_{j \rightarrow i}}
\label{eq:pi}
\end{equation}
\annotate[yshift=.25em]{above, left}{Pi}{probability of finding \\ \sffamily\footnotesize profile $i$}
\annotate[yshift=.5em]{above, right}{Pj}{probability of finding \\ \sffamily\footnotesize mutationally adjacent profile $j$}
\annotate[yshift=-1.2em]{below, left}{Pji}{probability of finding profile $i$ from profile $j$}


Where $a(x)$ is the set of score profiles adjacent to profile $x$:

\begin{equation}
    \eqnmarkbox[orange]{ai}{a(x)} = \{ y | \textit{Euclidean distance between x and y }= 1 \}
\label{eq:adj}
\end{equation}

The probability of transitioning from score profile $i$ to profile $j$ is given by the following equation:

\vspace{1.5em}

\begin{equation}
   \eqnmarkbox[red]{Pij}{P_{i \rightarrow j}} = \eqnmarkbox[teal]{survival}{P_{survival}(i, S, G, pop)} * \eqnmarkbox[blue]{mu}{\mu}
\label{eq:p_transition}
\end{equation}

\annotate[yshift=-1em]{below, right}{Pij}{probability of finding profile $j$ from profile $i$}
\annotate[yshift=1em]{above, right}{survival}{probability of profile \\ \sffamily\footnotesize $i$ surviving $G$ generations}
\annotate[yshift=.1em]{below, right}{mu}{mutation rate}

\vspace{2em}

Whereas the probability of score profile $i$ transitioning to itself (i.e. the probability of it remaining in the population for $G$ generations) is given by $P_{\text{survival}}$:

\vspace{1.2em}

\begin{equation}
    \eqnmarkbox[red]{Pij}{P_{i \rightarrow i}} = \eqnmarkbox[teal]{survival}{P_{survival}(i, S, G, pop)}
\label{eq:p_stay}
\end{equation}
\annotate[yshift=-.5em]{below, right}{Pij}{probability of profile $i$ staying in the population}
\annotate[yshift=.5em]{above, right}{survival}{probability of profile \\ \sffamily\footnotesize $i$ surviving $G$ generations}

\vspace{1em}

The contents of the population at time $t$ can be calculated using the following set union calculation. Formally, the population at time $t$ is a fuzzy set \citep{zadehFuzzySets1965}, where each member has a probability between 0 and 1 of actually being in the set:

\begin{equation}
    \eqnmarkbox[green]{pop}{pop_t} = \bigcup_{i\in \eqnmarkbox[green]{pop1}{pop_{t-1}}} \bigcup_{j\in \eqnmarkbox[orange]{ai}{a(i)}} \eqnmarkbox[red]{Pij}{(i, P_{j \rightarrow i})} 
\label{eq:pop_t}
\end{equation}

\vspace{2em}

\annotate[yshift=-.5em]{below, right}{Pij}{probability of profile $i$ being in set \\ \sffamily\footnotesize  is probability of finding $i$ from $j$}
\annotate[yshift=-.5em]{below, left}{ai}{profiles adjacent to $i$}
\annotate[yshift=-.5em]{below, left}{pop}{population at time $t$}

As discussed in the previous section, the set of Pareto-optimal score profiles selectable by lexicase selection is defined as:

\begin{equation}
    \eqnmarkbox[yellow]{f}{F} = \{x | \textit{x contains one value = L and D-1 values = 0} \}
\label{eq:fail_set}
\end{equation}

Finally, we can calculate the probability that lexicase selection fails to find any Pareto-optimal score profiles:

\begin{equation}
    P_{fail} = \prod_{i \in \eqnmarkbox[yellow]{f}{F}} \eqnmarkbox[violet]{Pi}{(1 - P_i)}
\label{eq:p_fail}
\end{equation}

\annotate[yshift=.1em]{above, right}{Pi}{probability of not finding $i$}

Because Equation \ref{eq:plex} requires the population to be a non-fuzzy set, this model can only be analyzed through running it stochastically. On each time step, we ``de-fuzzify'' the set $pop_t$ by creating a non-fuzzy set that includes or excludes each member with the appropriate probability. We implemented this stochastic model using Python 3.8.10 and C++. All code is open source and available at [REDACTED].

Prior work using the same model validated it against empirical experiments \cite{10.1145/3583133.3590714}. The probabilities of failing to find a Pareto-optimal solution predicted by the model closely matched those observed in a set of actual runs of lexicase selection. Thus, we feel confident that it accurately predicts the behavior of lexicase selection, and so use it here to predict the behavior of lexicase selection in a more complex and abstract scenario. Note that the reason the model is more efficient than experiments is because it uses the $G$ parameter to effectively skip over large portions of evolutionary time that must normally be spent traversing neutral regions of the fitness landscape.

\section{Results and Discussion}

\subsection{Analytic results}

Because our objectives are all maximally contradictory, any individual with a non-zero score on an objective will necessarily have negative scores on all other objectives. Thus, each individual with a non-zero score on an objective is effectively a specialist on that objective, and requires at least one selection event where that objective is chosen first in order to survive \citep{lacavaProbabilisticMultiObjectiveAnalysis2018}. Given this fact and the equations in Section \ref{sec:prior_math}, there is a clear mathematical limit on the parts of parameter space where lexicase selection can perform well on contradictory objectives. The lowest value of $P_{lex}$ that would allow a solution to reliably survive for $G$ generations is given by setting Equation \ref{eq:p_survival} equal to $t$, and solving for its $P_{lex}$ term:

\begin{equation}
1 - (1 - t^{\frac{1}{G}})^{\frac{1}{S}}    
\label{eq:p_needed}
\end{equation}

Because there are $D$ objectives, the probability of selecting a specific one first is $\frac{1}{D}$. For an alternative formulation of this reasoning that only considers one generation at a time, see \citep{lacavaProbabilisticMultiObjectiveAnalysis2018}. Combining this insight with equation \ref{eq:p_needed}, we can see that it is impossible for lexicase selection to find Pareto-optimal solutions to our problem unless the following inequality is satisfied:

\begin{equation}
1 - (1 - t^{\frac{1}{G}})^{\frac{1}{S}}  \leq \frac{1}{D}  
\label{eq:p_needed_vs_D}
\end{equation}

\begin{figure}
    \centering
    \includegraphics[width=\linewidth]{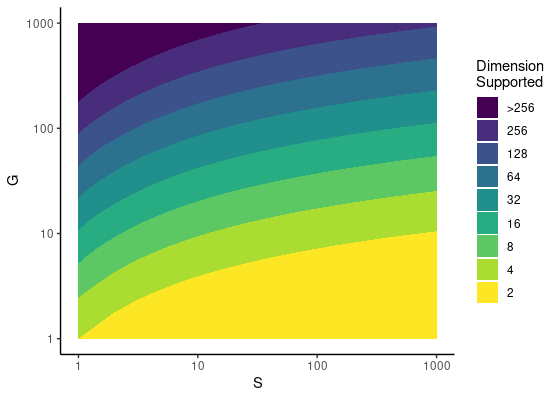}
    \caption{Visualization of the regions of parameter space where lexicase selection can find Pareto-optimal solutions to a problem with contradictory objectives, as given by Equation \ref{eq:p_needed_vs_D}, with $t=.5$. Colors indicate upper bounds on the dimension (i.e. number of objectives) that can be used for each combination of $S$ and $G$. Note that the x, y, and color axes are all on log scales.}
    \label{fig:G_vs_S_equation}
\end{figure}

Note that, although we are focusing here on a particularly extreme example problem that falls outside lexicase selection's traditional domain of application, this inequality has more general implications for lexicase selection. Specifically, it describes the limits on circumstances where a candidate solution can survive in the long term on the basis of performing well on a single objective. As most problems lexicase selection is used on have some aligned objectives, it is likely okay in most cases to choose parameters such that candidate solutions need to perform well on more than one solution. Nevertheless, choosing $D$, $G$, and $S$ such that this inequality is not satisfied may cut off access to certain paths through the search space. Users of lexicase selection should think critically about this fact in the context of their objectives when selecting parameters.

\begin{figure*}
    \centering
    \includegraphics[width=\textwidth]{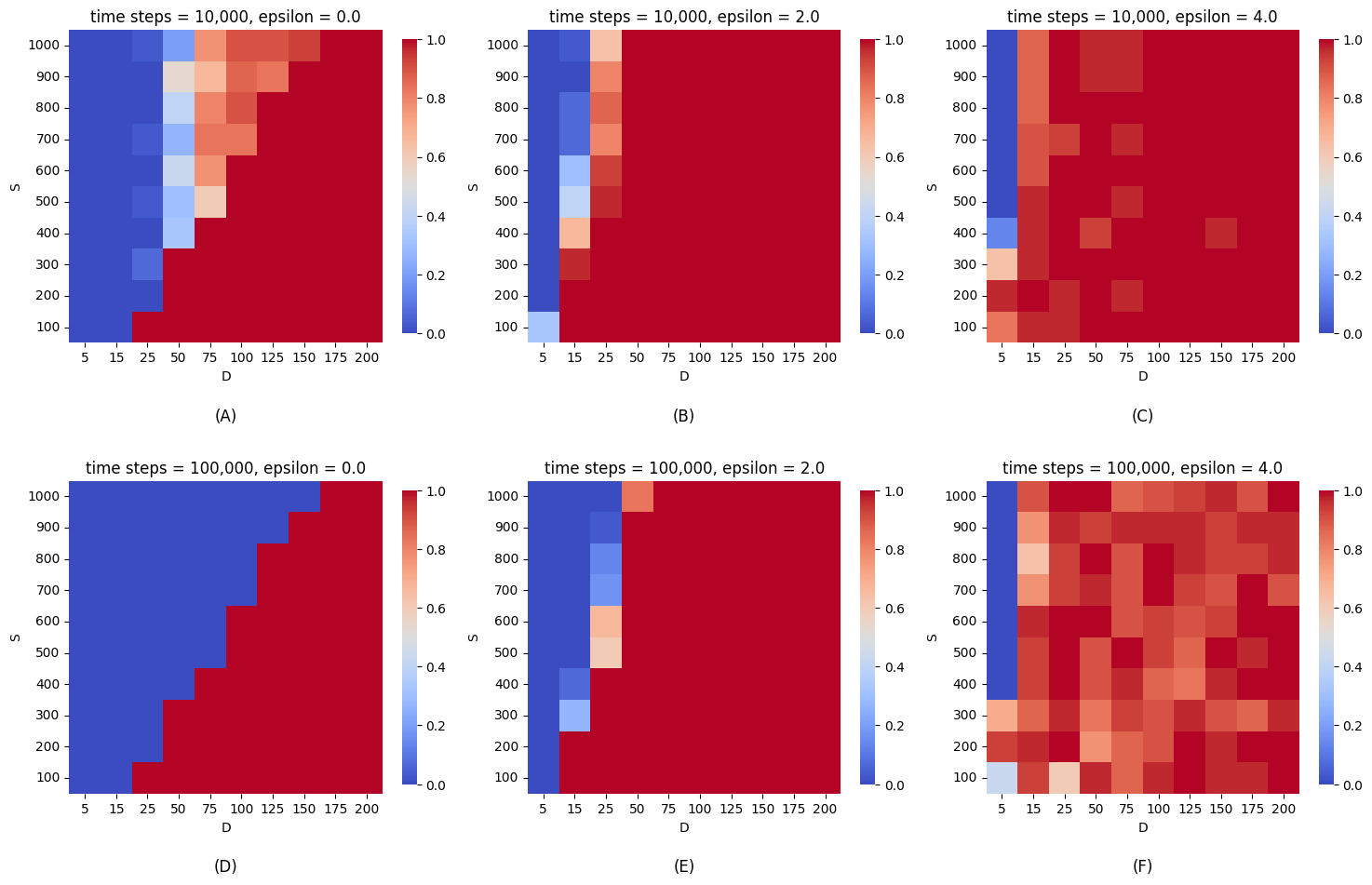}
    \caption{Probability that lexicase selection fails to find a Pareto-optimal solution over various values of $S$, $D$, and $\epsilon$. In figures (A), (B) and (C), the algorithm was run for 10,000 time steps while in figures (D), (E) and (F) it was run for 100,000 time steps (n = 30 per cell, $\mu$=.01, $G$=500).}
    \label{fig:all}
\end{figure*}

These mathematical results are highly consistent with empirical results obtained by Hernandez et. al \citep{Hernandez2022}. They used population size 512 and ran for 50,000 generations. On average, lexicase selection maintained a Pareto front containing solutions with optimal scores on approximately 40 out of 100 objectives (non-dominated sorting was able to maintain approximately 90 on average). Plugging Hernandez et. al's parameters into Equation \ref{eq:p_survival}, we can relate the probability of a Pareto-optimal solution surviving to the number of specialists on other objectives currently in the population:

\begin{equation}
    (1 - (1 - 1/D)^{512})^{50000}
\label{eq:p_survival_hernandez}
\end{equation}

This equation is equal to .5 when $D\approx46$, very close to the average number of contradictory objectives that lexicase selection simultaneously optimized in this configuration. However, this equation only gives the probability of a single specialist surviving 50,000 generations when competing against $D$ other types of specialist. The probability of $D$ unique specialists surviving is given by: 

\begin{equation}
    ((1 - (1 - 1/D)^{512})^{50000})^D
\label{eq:p_survival_hernandez_underestimate}
\end{equation}

This equation is a slight underestimate of the number of unique specialists that should survive in Hernandez et. al's work, as it neglects to account for the fact that there are more than D objectives to potentially specialize on. Nevertheless, this equation is equal to .5 when $D\approx35$, which is very close to the value Hernandez et. al's distribution of results is centered on. This observation provides further support for the validity of our mathematical analysis, and for Hernandez et. al's suggestion that these results were driven by population size.

This re-analysis of Hernandez et. al's contradictory objectives experiment sheds light on a key difference between lexicase selection and non-dominated sorting: the scaling relationship between maximum maintainable Pareto front size and population size. Non-dominated sorting clearly has an advantage in this regard, as the size of the Pareto front it can maintain scales linearly with population size. However, La Cava and Moore's results suggest that lexicase selection may have an advantage with regard to its ability to efficiently search a high-dimensional space \citep{lacavaAnalysislexicaseSelection2018a}.

\subsection{Stochastic Modelling Results}
\label{sec:math}

The modelling results support our analytic conclusion that satisfying Equation \ref{eq:p_needed_vs_D} is essential if lexicase selection is to discover the Pareto-optimal solutions to this problem. As predicted, standard lexicase selection ($\epsilon=0$) never succeeds unless $S$ is high enough relative to $D$ and $G$ (see Figures \ref{fig:all}-A and \ref{fig:all}-D). If $S$ is high enough, success appears to depend entirely on whether search is given enough time to find Pareto-optimal values. Higher values of $D$ appear to require more time to find solutions (compare Figures \ref{fig:all}-A and \ref{fig:all}-D). This difference is to be expected, as our mutation rate is per-genome, so genotypes with bigger dimensions receive fewer mutations per value, thus taking longer to traverse the fitness landscape. Previous work using the same model tested this assumption by changing $\mu$ in proportion to $D$ \cite{10.1145/3583133.3590714}. Under those circumstances, lexicase selection demonstrated better performance in finding a Pareto-optimal solution within the time limit at higher values of $D$ than at lower values of $D$. This observation suggests that lower dimensions require a higher per-site mutation rate to reach an optimum as swiftly as higher dimensions. 
Ultimately, we conclude that as long as the values of $S$, $D$ and $G$ satisfy equation \ref{eq:p_needed_vs_D}, standard lexicase selection can successfully solve many-objective optimization problems with contradictory objectives. 

$\epsilon$-lexicase selection appears to be slightly less robust to the presence of many contradictory objectives. As $\epsilon$ increases, the ratio of $S$ to $D$ required to find Pareto-optimal solutions appears to increase (see Figures \ref{fig:all}-B-C and \ref{fig:all}-E-F). Again we see that there is a region of parameter space where finding a Pareto-optimal solution appears to be impossible. As in our results for $\epsilon=0$, some conditions along the edge of this region appear to require more time to find a Pareto-optimal solution (compare Figures \ref{fig:all}-B and \ref{fig:all}-E). As $\epsilon$ gets larger, the size of this intractable region of parameter space appears to grow. One important caveat, however, is that making $\epsilon$ very large results in stochastically finding the optimum even at very high values of $D$ and low values of $S$ (see Figure \ref{fig:all}-C), presumably due to drift.


\begin{figure}
    \centering
    \includegraphics[width=\linewidth]{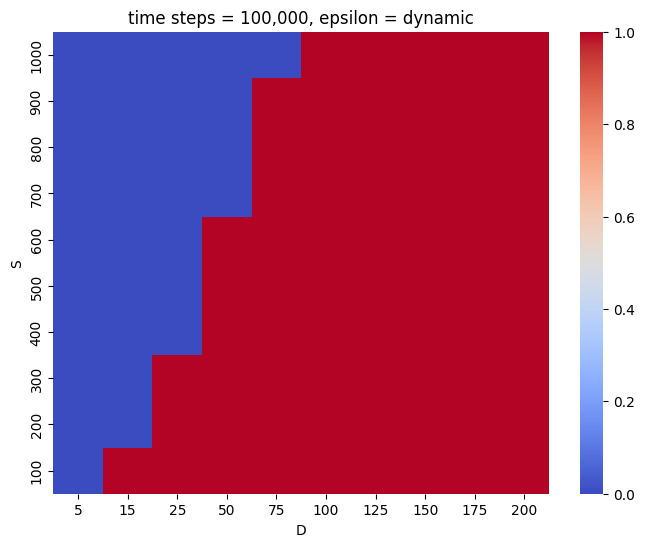}
    \caption{Probability that $\epsilon$-lexicase selection fails to find a Pareto-optimal solution over various values of S and D when $\epsilon$ changes according to equation \ref{eq:medianepsilon}. The algorithm was run for 100,000 time steps (n = 30 per cell, $\mu$=.01, $G$=500)}
    \label{fig:dynamicepsilon}
\end{figure}

While these results are consistent with prior empirical observations showing that epsilon lexicase selection struggles with extremely contradictory objectives \citep{shahbandegan_untangling_2022, 10.1145/3583133.3590714}, they are surprising given the success that $\epsilon$-lexicase selection has demonstrated on multi-objective problems \cite{lacavaProbabilisticMultiObjectiveAnalysis2018, lacavaAnalysislexicaseSelection2018a}. One possible explanation is that we have made a simplifying assumption here by holding $\epsilon$ constant. This assumption was also made in \citep{shahbandegan_untangling_2022, 10.1145/3583133.3590714}. In contrast, \cite{lacavaProbabilisticMultiObjectiveAnalysis2018, lacavaAnalysislexicaseSelection2018a} used a variable $\epsilon$ value. To address this disparity, we conducted additional stochastic modelling experiments where $\epsilon$ was dynamically adjusted in each generation based on equation \ref{eq:medianepsilon}, as in \cite{lacavaProbabilisticMultiObjectiveAnalysis2018, lacavaAnalysislexicaseSelection2018a}. The results of this investigation are illustrated in Figure \ref{fig:dynamicepsilon}. When we adjust $\epsilon$ each generation, lexicase selection finds more Pareto-optimal solutions than when we hold $\epsilon$ constant at 2 or 4. The one downside of a dynamic epsilon appears to be that it cannot find Pareto-optimal solutions at high values of $D$ via drift (as was possible when $\epsilon=4$). However, as drift is not a particularly reliable way to find solutions, this trade-off is likely worthwhile. 

These results suggest: 1) setting $\epsilon$ using median absolute deviation substantially increases $\epsilon$-lexicase selection's ability to optimize many contradictory objectives simultaneously, and 2) even with this modification, $\epsilon$-lexicase selection is slightly less robust to the presence of many contradictory objectives than standard lexicase selection is. To understand why, we turn to reachability analysis.

\subsection{Reachability results}

One takeaway from \ref{fig:all} is that some failures of lexicase and $\epsilon$-lexicase selection to find Pareto-optimal solutions are the result of insufficient search time rather than the solutions truly being impossible to find. This observation raises an important question: in which conditions are Pareto-optimal solutions impossible to find and in which conditions is finding them just very slow? To answer this question, we performed reachability analyses using the methodology described in \cite{dolson2023reachability}. More specifically, in experiments where lexicase selection failed to find Pareto-optimal solutions, we recorded the final population of individuals. We then explored the state-space of population compositions that could theoretically be reached from that point to see if it contained any Pareto-optimal solutions. Because the state-space of lexicase selection is large, there were three possible results from this analysis: 1) \textbf{reachable}, meaning that the region of state-space we explored contained a Pareto-optimal solution, 2) \textbf{unreachable}, meaning that we were able to explore the entire region of state-space accessible from the final population and it did not contain any Pareto-optimal solutions, or 3) \textbf{indeterminate}, meaning that the reachable region of state space was too large to explore in its entirety, but the portion that we visited did not contain any Pareto-optimal solutions. 

As expected, the reachability analyses showed that when $\epsilon = 0$ and we fail to find an optimal solution after 100,000 time steps, it is always because optimal solutions are unreachable. These are the conditions where $S$ is too small relative to $G$ and $D$; the population gets stuck at the starting point (where are values in the genotypes are 0) and is unable to escape that state. Intuitively, this result makes sense because any mutation producing a score greater than 0 necessarily causes all other scores to be less than 0. The new solution is thus a specialist on a single objective which, as our mathematical analysis tells us, cannot reliably survive under such conditions. As a result, even given infinite time, lexicase selection cannot find Pareto-optimal values under these conditions. 

As $\epsilon$ gets bigger, however, it is less intuitively clear what is the cause of failures to find Pareto-optimal solutions. Thus, our reachability analysis becomes more important. Again, we find that there are cases when Pareto-optimal solutions are not reachable even given infinite time. Unexpectedly, these cases occur mostly when population size is large. At smaller population sizes, the reachability of Pareto-optimal solutions is generally indeterminate. Intuitively, we would expect a higher population size to support the long-term survival of more specialists, supporting evolution towards Pareto-optimal solutions.

 \begin{figure}
    \includegraphics[width =.8\linewidth]{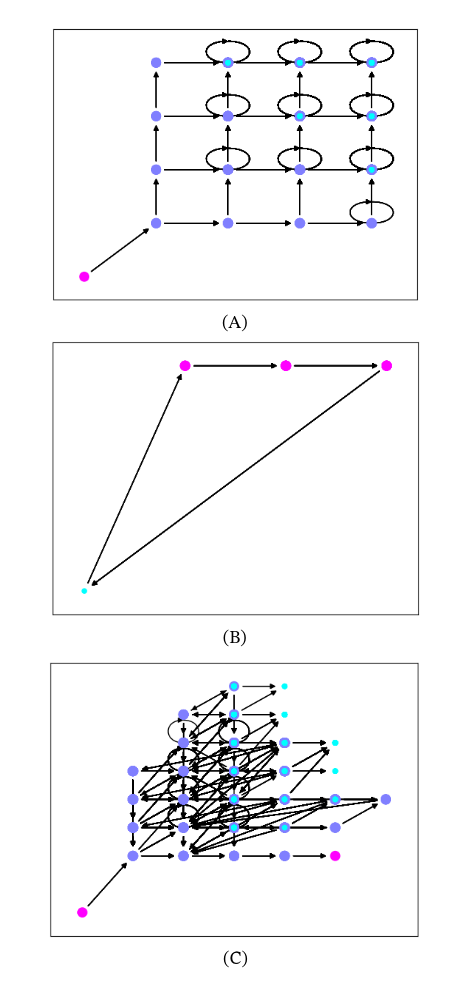}
    \caption{State space explored in a down-scaled reachability analysis graph for $D=5$, $S=30$, $G=50$. A) $\epsilon$=0, B) $\epsilon$=1, C) $\epsilon$=2. Position along the x axis indicates the number of unique genotypes. Position along the y axis correlates with the distance of genotype values from zero. To make the structure visually clear, nodes with similar properties in these regards are plotted on top of each other. Cyan nodes were discovered but not fully explored in the reachability analysis.}
    \label{fig:cags}
\end{figure}

To understand why $\epsilon$-lexicase has such different reachability properties from standard lexicase selection, we conduct a reachability analysis on a scaled-down version of our landscape ($S=30$, $D=5$, $G=50$) starting from the same point our modelling runs started (a population containing only the genotype with zeroes in all positions). We explore the 1000 population compositions that are most reachable from this point with $\epsilon$ values of 0, 1, and 2.

We find strikingly different behavior across these conditions (see Figure \ref{fig:cags}). At $\epsilon=0$, evolution climbs a strict gradient with no opportunity to backtrack. At $\epsilon=1$, evolution is stuck in a loop from which it cannot escape. At $\epsilon=2$, the state space is complex. A sink node exists, indicating that there is potential for evolution to get trapped in a state from which it cannot escape. The rest of the graph is all part of a single strongly-connected component. While there is potential to climb towards genotypes with higher scores on a single objective, it is also possible to backtrack from those population compositions. If there is a gradient for evolution to climb, it is much weaker than the one in standard lexicase selection.

While these results do not provide conclusive explanations for why $\epsilon$-lexicase struggles more with contradictory objectives than standard lexicase selection does, they provide qualitative intuition. Clearly, higher values of $\epsilon$ complicate the search process and create potential for it to get stuck.

\section{Conclusions}

We have identified a region of parameter space where lexicase selection is unable to find Pareto-optimal solutions to a problem with many contradictory objectives. This finding is supported by analytic and modelling results, and it is consistent with prior empirical work. In $\epsilon$-lexicase selection, the size of this region of parameter space increases with $\epsilon$. Adjusting $\epsilon$ based on the current population reduces the size of this region, but it remains larger than under standard lexicase selection.  
These results provide robust, theory-backed guidelines on the selection of parameters for lexicase selection on this style of problem. They also provide support for adjusting $\epsilon$ based on median absolute deviation.

Outside of this region of parameter space, our analytic, modeling, and prior empirical results all suggest that lexicase and $\epsilon$-lexicase selection are able to perform well on this problem. This observation is surprising, because this class of problems is very different from those for which lexicase selection was designed. It also lends further support to the idea that lexicase selection has potential as a massive objective optimization algorithm. Moreover, as the number of objectives increases, the chances that they will truly all be maximally contradictory seem low, bolstering our confidence that these results represent a worst-case scenario.

While this work has provided good insights into the behavior of lexicase and $\epsilon$-lexicase selection under contradictory objectives, it has also raised many new questions. In the future, we will conduct more extensive reachability analyses to understand why $\epsilon$-lexicase selection failed to find Pareto-optimal solutions under some conditions where  standard lexicase succeeded. In addition, we plan to investigate the impact of the intensity of conflict among the objectives. Prior research suggests that $\epsilon$-lexicase may be more able to handle weakly contradictory objectives than it is to handle maximally contradictory objectives \citep{shahbandegan_untangling_2022}. By quantifying the amount of contradiction that lexicase and $\epsilon$-lexicase selection can tolerate, we can predict which many-objective optimization problems they are appropriate for. Lastly, we hope the community will test lexicase selection on a wider variety of many-objective optimization problems to better understand its potential in this domain.

\begin{acks}
We would like to thank the ECODE lab

\end{acks}

\bibliographystyle{ACM-Reference-Format}
\bibliography{sample-base}










\end{document}